\documentclass{esann}
\usepackage[dvips]{graphicx}
\usepackage[latin1]{inputenc}
\usepackage{amsmath}
\usepackage{amsfonts}
\usepackage{setspace}
\usepackage{enumerate}
\usepackage{xytree}
\usepackage{color}
\usepackage{hyperref}
\usepackage{listings}
\usepackage{hyperref}
\usepackage{framed}
\usepackage{pstricks}
\usepackage{pst-node}
\usepackage{pst-tree}
\usepackage{natbib}
\usepackage{dirtytalk}

\begin{document}
%style file for ESANN manuscripts
\title{Fine-grained Event Learning of Human-Object Interaction with LSTM-CRF}

%***********************************************************************
% AUTHORS INFORMATION AREA
%***********************************************************************
\author{Tuan Do and James Pustejovsky
%
% Optional short acknowledgment: remove next line if non-needed
\thanks{This work is supported by a contract with the US Defense Advanced Research Projects Agency (DARPA), Contract W911NF-15-C-0238.
 Approved for Public Release, Distribution Unlimited. The views expressed are those of the authors and do not reflect the official policy or position of the Department of Defense or the U.S. Government.  We would like to thank Nikhil Krishnaswamy and
Keigh Rim for their discussion and input on this topic.  All errors and mistakes are, of course, the responsibilities of the authors..}
%
% DO NOT MODIFY THE FOLLOWING '\vspace' ARGUMENT
\vspace{.3cm}\\
%
% Addresses and institutions (remove "1- " in case of a single institution)
Brandeis University - Department of Computer Science \\
Waltham, Massachusetts - United States of America
%
% Remove the next three lines in case of a single institution
}
%***********************************************************************
% END OF AUTHORS INFORMATION AREA
%***********************************************************************

\maketitle

\begin{abstract}

Event learning is one of the most important problems in AI. However, notwithstanding significant research efforts, it is still a very complex task, especially when the events involve the interaction of humans or agents with other objects, as it  requires modeling human kinematics and object movements. This study proposes a methodology for learning  complex human-object interaction (HOI) events, involving the recording, annotation and classification of event interactions. For annotation, we allow multiple interpretations of a motion capture by slicing over its temporal span; for classification, we use Long-Short Term Memory (LSTM) sequential models with Conditional Randon Field (CRF) for constraints of outputs. Using a setup involving captures of human-object interaction as three dimensional inputs, we argue that this approach could be used for event types involving complex spatio-temporal dynamics. \iffalse The results showed that the proposed method achieved high improvement over baseline on our setup using three dimensional data captured. \fi
\end{abstract}

\section{Introduction}

The study of events has long involved many disciplines, including philosophy, cognitive psychology, linguistics, computer science, and AI. The Gestalt school of philosophy characterized events as whole processes that emerge from the relations between their components. Cognitive psychologists, such as Tulving \cite{tulving1983elements}, recognized the importance of events by postulating a separate cognitive process called \textit{episodic memory}. The representation of events in natural language has been studied from many different approaches, from formal logic and AI \cite{Allen:1984}, and frames \cite{Fillmore75}, to computational linguistics \cite{TimeML-LRE:2011}. 
%addressed in the same framework with temporal expressions in TimeML %\cite{TimeML:2003}, as that allows reasoning on temporal order and %%viewpoint, 
Combining perspectives from computer science, logic, and linguistics, some recent work  suggests that events can be effectively modeled  as 
 programs within a dynamic logic (DITL) \cite{pustejovsky2011qualitative}, enabling computer simulations of linguistic expressions  \cite{pustejovsky2014generating}.
% that enables computer simulation.

In computer science, there is little consensus about how events should be modeled for learning. They can be represented atomically, i.e., entire events are predicted in a classification manner \cite{shahroudy2016ntu}, or as combinations of more primitive actions \cite{veeraraghavan2007learning}, i.e., complex event types are learned based on recognition of combined primitive actions. For the former type of event representation, there are quantitative approaches based on low-level pixel features such as in \cite{le2011learning} and qualitative approaches such as induction from relational states among event participants \cite{dubba2015learning}. For the latter approach, systems such as \cite{hoogs2008video}, use state transition graphical models such as Dynamic Bayesian Networks (DBN). 

While learning events as a whole works best for human motion signatures such as \textit{running}, \textit{sitting} etc., it poses a problem for event types that require distinctions in spatio-temporal relationships between objects. As pointed in \cite{dubba2015learning}, it is also difficult to model events as  strict orderings of subevents, especially when there are \textit{overlapping} or \textit{during} relations between them. Moreover, if the purpose of event learning is to facilitate communication and interaction between human and computational agents, such as robots, to achieve some common goals, these agents need to keep track of multiple events at the same time, involving themselves, other humans, as well as the surrounding environment. From a practical point of view, this calls for a  finer-grained treatment of event modeling.

\iffalse Take the event \say{A person rolls A past B} as an example. Zooming into different parts of the event, we see different things. At the beginning, the performer reaches his hand to A, A starts to roll closer to B till some point of time that we can claim that A moves past B. A might continue rolling, with or without force from the performer's hand. \fi

\iffalse
For example, imagine a scenario in which a human agent takes a \textit{pedagogic role} and a robotic agent needs to learn to perform gradually more complex activities. 
\fi

It is also the case that
%It is noted that the motivation for that treatment
a fine-grained analysis of events is strongly supported from a theoretical point of view. For example, it has long been known that event classification needs to take into account what is called \textit{extra-verbal factors}. Event types should not be semantically defined only by a base verbal expression, such as \textit{running} or \textit{walking}, but  need to incorporate other components of the expression compositionally, such as objects and adjuncts, which can change the event type of the overall verb phrase or sentence \cite{Pustejovsky1995}. 

Motivated by those arguments, we suggest a different approach to event learning. Instead of treating events as whole, or as programs of subevents, we allow multiple interpretations of a motion capture by slicing over its temporal span and give a separate annotation for each slice. 

In particular, we use an event capture and annotation tool called ECAT \cite{do2016ECAT}, which employs Microsoft Kinect\circledR\ to capture sessions of performers interacting with two types of objects, a cube (which can be slid on a flat surface) and a cylinder (which can be rolled). Objects are tracked using markers fixed to their sides facing the camera. They are then projected into three dimensional space using depth of field. Performers are tracked using the Kinect API, which provides three dimensional inputs of a performer's joint points (e.g., wrist, palm, shoulder). Sessions are first sliced, and each slice is annotated with a textual description using our event language. Our sequence learning algorithms (LSTM-CRF) will input sequences of feature vectors and output a representation of an event. 

The main contributions of our study are twofolds. Firstly we created a framework for event recording and annotation that takes into account their temporal dynamics, i.e., different interpretations of events on different temporal spans. Applying a flavor of the popular sequential learning method LSTM that accommodates to output constraints, we achieved good performance in our human-object interaction setup.

\section{Learning Framework}

We used three dimensional coordinates of bodies tracked by Kinect\circledR SDK to model human kinematics. Only joint points on the upper body of performers (13 joint points) are kept and concatenated, because their lower parts are occluded when they interact with objects on top of a table. In addition, for each marker detected (using Glyph detection algorithm\cite{glyph}), we generate 12 features (4 3-D corners). Features of objects are concatenated into a vector of a fixed size (there is always one performer and two objects tracked). A sample in the dataset consists of a sequence of fixed length of feature vectors. Its label is mapped from the textual annotation into an output structure (Subject, Object, Locative, Verb, Preposition). We will call this output structure $(Y_1,\ldots,Y_5)$.

\subsection{LSTM}

LSTM is a flavor of deep Recursive neural network (RNN) that has generally solved the problem of \say{vanishing gradients} in traditional RNN learning \cite{hochreiter1997long,schmidhuber2015deep} and has found their application in a wide range of problems involving sequential learning, such as hand written recognition, speech recognition, gesture recognition, etc. 

\begin{figure}[!htbp]
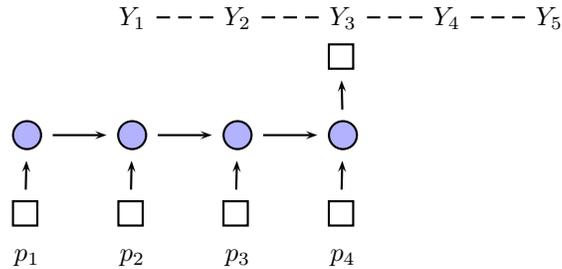

\begin{center}
$
\psmatrix[colsep=1cm,rowsep=0.1cm]
& [name=Y_1] Y_1 & [name=Y_2] Y_2 & [name=Y_3] Y_3 & [name=Y_4] Y_4 & [name=Y_5] Y_5 \\
& & & [name=O, mnode=f]\\
 \\
[name=H_1, mnode=C,radius=0.2,fillstyle=solid,fillcolor=blue!30] & [name=H_2, mnode=C,radius=0.2,fillstyle=solid,fillcolor=blue!30] & [name=H_3, mnode=C,radius=0.2,fillstyle=solid,fillcolor=blue!30] & [name=H_4, mnode=C,radius=0.2,fillstyle=solid,fillcolor=blue!30] \\
\\
[name=I_1,mnode=f]  & [name=I_2,mnode=f] & [name=I_3,mnode=f] & [name=I_4,mnode=f]\\
p_1 & p_2 & p_3 & p_4
\ncline[nodesep=4pt,linestyle=dashed]{-}{Y_4}{Y_5}
\ncline[nodesep=4pt,linestyle=dashed]{-}{Y_3}{Y_4}
\ncline[nodesep=4pt,linestyle=dashed]{-}{Y_2}{Y_3}
\ncline[nodesep=4pt,linestyle=dashed]{-}{Y_1}{Y_2}
\ncline[nodesep=4pt]{->}{H_4}{O}
\ncline[nodesep=4pt]{->}{H_3}{H_4}
\ncline[nodesep=4pt]{->}{H_2}{H_3}
\ncline[nodesep=4pt]{->}{H_1}{H_2}
\ncline[nodesep=4pt]{->}{I_1}{H_1}
\ncline[nodesep=4pt]{->}{I_2}{H_2}
\ncline[nodesep=4pt]{->}{I_3}{H_3}
\ncline[nodesep=4pt]{->}{I_4}{H_4}
\endpsmatrix
$
\end{center}
\caption{LSTM model with possible constraints of outputs with CRF. CRF layer is represented as dashed links among predicted labels.}
\end{figure}

We will not describe in detail here our LSTM implementation, as we provide  online access to the code and approach \footnote{https://github.com/tuandnvn/ecat\_learning}. Briefly, however, the model passes each feature vector through a linear layer before feeding each sequence into an LSTM. Each label $Y_i$ requires a separate LSTM cell, $X_i$. Depending on whether we predict each label independently and combine them for final prediction, or we predict on the basis of the sum of outputs from the last layer, two variants are considered, correspondingly LSTM-I and LSTM-W.

\subsection{CRF}

% Shadow box
\newcommand{\shb}[1]{[name=#1,mnode=r] \psframebox[shadow=true, framearc = 0.25,framesep=0.1]{#1}}

\begin{figure}[!htbp]
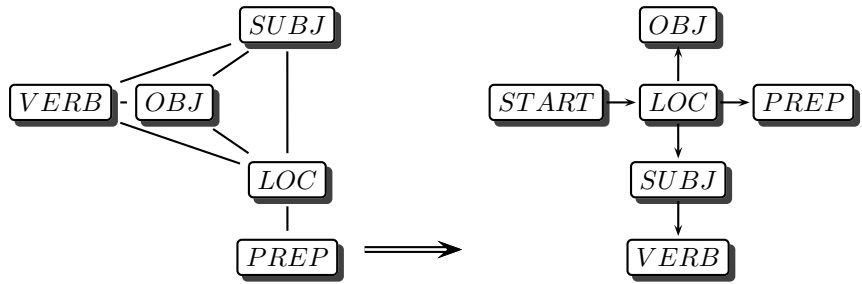

\hspace*{-2cm}
\centering
$
\psmatrix[colsep=2cm]
[name=original-CRF]
\psmatrix[colsep=0.3cm,rowsep=0.5cm]
&&  \shb{SUBJ} \\
\shb{VERB} & \shb{OBJ} \\
&& \shb{LOC}\\
&& \shb{PREP}
\ncline[nodesep=3pt]{SUBJ}{OBJ}
\ncline[nodesep=3pt]{OBJ}{LOC}
\ncline[nodesep=3pt]{SUBJ}{LOC}
\ncline[nodesep=3pt]{SUBJ}{VERB}
\ncline[nodesep=3pt]{OBJ}{VERB}
\ncline[nodesep=3pt]{LOC}{VERB}
\ncline[nodesep=3pt]{LOC}{PREP}
\endpsmatrix
&
[name=tree-CRF]
\psmatrix[colsep=0.3cm,rowsep=0.5cm]
& \shb{OBJ} \\
\shb{START} & \shb{LOC} & \shb{PREP}  \\
& \shb{SUBJ} \\
& \shb{VERB}
\ncline[nodesep=1pt]{->}{START}{LOC}
\ncline[nodesep=1pt]{->}{LOC}{SUBJ}
\ncline[nodesep=1pt]{->}{LOC}{OBJ}
\ncline[nodesep=1pt]{->}{LOC}{PREP}
\ncline[nodesep=1pt]{->}{SUBJ}{VERB}
\endpsmatrix
\ncline[doubleline=true,nodesep=10pt]{->}{original-CRF}{tree-CRF}
\endpsmatrix
$
\caption{Reformation from general {\bf CRF} (left) to {\bf Tree-CRF} (right)}
\label{fig:crf}
\end{figure}

CRF has been used extensively to learn structured output as it allows specification of constraints of output labels\cite{sutton2006introduction}. In this model we wish to constrain the outputs so that: one object (Performer or the other objects) is not allowed to fill  two different syntactic slots; when there is no verb, all the other slots should be None; locative and preposition are dependent, because if locative is None, preposition must also be None and vice versa. The edges between nodes on the left side of Figure~\ref{fig:crf} show the dependencies on output labels that we wish to model.

However, training and classifying using a full CRF model would be more difficult, especially when implemented  with a neural network architecture.  We modified the model into a tree-CRF structure (right side of Figure~\ref{fig:crf}) to make the model learnable using dynamic programming. The complexity of the algorithm reduced from $O(n^5)$ to $O(n^2) * 5)$ where $n$ is the size of our vocabulary. The learning problem is thereby changed to learning the weights along the edges on the tree-CRF, for example, $P\_locative\_preposition$ (together with parameters of LSTM). The directionality of edges is the forward direction of the message passing algorithm used for learning (and in reverse, for testing using the backward direction).

\subsection{LSTM-CRF}

LSTM-CRF is a natural extension of LSTM applied for constrained outputs. For instance it is used for named entities recognition task to model constraints on BIO labels\cite{huang2015bidirectional}. To put CRF learning on top of LSTM-W, we modify the term $t$ (the term before softmax) produced by outputs of LSTM as followings.

\begin{align*}
t(l,s,o,p,v) &= t_l + t_s + t_o + t_p + t_v \text{\vspace{20mm} \textbf{     Original LSTM-W}} \\
t(l,s,o,p,v) &= t_l + t_s + t_o + t_p + t_v \text{\vspace{20mm} \textbf{     Modified}}\\
&+ P_{start\_l} + P_{ls} + P_{lo} + P_{lp} + P_{sv}
\end{align*}

\noindent where $l$, $s$, $o$, $p$, $v$ stand for Locative, Subject, Object, Preposition and Verb respectively. 

In training, $softmax$ is calculated for a predicted label combination, namely $(l', s', o', p', v')$ as below. We can calculate the $log$ of $sum$ using message passing over the tree nodes of the CRF tree. We use cross entropy between predicted distribution and correct output as the $cost$ in training.

\begin{align*}
softmax &= exp[ t (l',s',o',p',v') -  log [\sum_{l}\sum_{s}\sum_{o}\sum_{p}\sum_{v} exp(t (l,s,o,p,v)) ] ] \\
&=  exp[ t (l',s',o',p',v')
-  log [ \sum_{l} exp(t_l + P_{start\_l})
[\sum_{s} exp(t_s + P_{ls}) \\ 
& \sum_{v} exp(t_v + P_{sv})]
[\sum_{o} exp(t_o + P_{lo})] [\sum_{p} exp(t_p + P_{lp})]]
\end{align*}

\noindent In evaluation, a similar message passing algorithm is used, but instead of $log\_sum$, we use $max$ to calculate the probabilities and $argmax$ to keep track of the best combination.

\section{Experiments}

\subsection{Event Capture and Annotation}

To demonstrate our model's capability to learn the \textit{spatio-temporal} dynamics of object interactions in events, we use a collection of four action types: \textit{push}, \textit{pull}, \textit{slide}, and \textit{roll}, along with  three different \textit{spatial} prepositions used for space configurations between objects, namely \textit{toward} (when the trajectory of a moving object is straightly lined up with a destination static object and makes it closer to that target), \textit{away from} (makes it further from that object) and \textit{past} (moving object getting closer to static object then further again).

Afterwards, for each session, we sliced the events into short segments of 20 frames. Two annotators were assigned to watch and annotate them (segments can be played back). To speed up annotation, only event types related to original captured types are shown for selection. For instance, if  the event type of the captured session is \say{The performer pushes A toward B}, other available event types are \say{The performer pushes A}, \say{A slides toward B} or \say{None}.  

\subsection{Classification Results}

Our LSTM models have one hidden layer of 200 features. Two methods are used to combat over-fitting: (i) dropout in LSTM cell with probability of 0.8, and (ii) gradient clipping by a global norm. The network is trained with mini-batch gradient descent optimization for 200 epochs on the Tensorflow library. 

Most frequent label tagging is used as the baseline for this study: that is, we simply predict any sample with the most frequent tuple seen in the training corpus.

Captured sessions are split into training and testing sets on the proportion of 60/40, i.e., 18 sessions of training and 12 sessions for testing for each event type. That gives a total of 2680 training samples and 1840 testing samples. Precisions reported are averaged over 5 runs (each run is obtained with a random initialization). Breakdowns of precision for each label show that verb precision is the lowest, with high confusion between the following pairs: \textit{push} vs \textit{roll}, \textit{pull} vs \textit{roll}, and \textit{slide} vs \textit{roll}. It is likely because of poor tracking result when objects are rolled. In fact, the capture tool could not recognize objects in many frames when objects roll fast, and it compensated by using interpolation. Improvement on tracking of objects, however, is not the target of our study.

\begin{table}[!ht]
\parbox{.45\linewidth}{
\scriptsize
%\caption{Evaluation}
\label{table:headings}
\centering
\begin{tabular}{|l|c|}
\hline
Model & Precision \\
\hline
Baseline & 6\% \\
\hline
LSTM-I & 38\% \\
\hline
LSTM-W & 39\% \\
\hline
LSTM-CRF & 43\% \\
\hline
\end{tabular}
\vspace{3mm}
\caption{Evaluation}
}
\hfill
\parbox{.45\linewidth}{
\scriptsize
%\caption{Label precision breakdown}
\label{table:headings}
\centering
\begin{tabular}{|l|c|}
\hline
Label & Precision \\
\hline
Subject & 86\% \\
\hline
Object & 87\% \\
\hline
Locative & 73\% \\
\hline
Verb & 68\% \\
\hline
Preposition & 72\% \\
\hline
\end{tabular}
\caption{Label precision breakdown}
}
\end{table}

\noindent We observed significant improvement of learning using LSTM over baseline, especially when it is coupled with CRF. Moreover, we also observed a reduction of invalid outputs from 20\% to 3\% when CRF is used.  We consider these  results to be quite good, particularly since the sequential learning model we used is both simple and fast, and we did not employ any feature engineering method.

\section{Conclusion and Future Directions}

In this paper, we have demonstrated a methodology that provides  reasonable learning results for a set of complex event types. We hope that our study will provide a starting point for further investigations into fine-grained event learning.
Currently our learning method requires a fix number of objects in inputs, which could be overcome by incrementally adding object features into a fix size feature vector, possibly by using a recursive neural network. Regarding our annotation framework, a natural extension is that spans of different lengths could be annotated with appropriate re-sampling methods. We  leave these as some of our future research topics.

We are currently applying this learning pipeline (simple interval annotation + sequential learning with constraint outputs) on a large movie dataset that have event annotations. We are looking to publish our results in the near future.

\begin{footnotesize}

% IF YOU USE BIBTEX,
% - DELETE THE TEXT BETWEEN THE TWO ABOVE DASHED LINES
% - UNCOMMENT THE NEXT TWO LINES AND REPLACE 'Name_Of_Your_BibFile'

\setlength{\bibsep}{1pt plus 0.5ex}

\bibliographystyle{unsrt}
\bibliography{References}

\end{footnotesize}

\end{document}